\documentclass[11pt]{article}

\usepackage{microtype} %
\usepackage{booktabs}  %
\usepackage{url}  %

\usepackage{amsmath}
\usepackage{amsthm}

\usepackage{amssymb}
\usepackage{mathtools}
\usepackage{amsthm}
\usepackage{multirow}

\usepackage[utf8]{inputenc} %
\usepackage[T1]{fontenc}    %
\usepackage{url}            %
\usepackage{booktabs}       %
\usepackage{amsfonts}       %
\usepackage{nicefrac}       %
\usepackage{microtype}      %
\usepackage{xcolor}         %
\usepackage{acronym}
\usepackage{amsmath}
\usepackage{graphicx}

\acrodef{NAS}{Neural Architecture Search}
\acrodef{HPO}{Hyperparameter Optimization}
\acrodef{ML}{Machine Learning}
\acrodef{DL}{Deep Learning}
\acrodef{NN}{Neural Network}
\acrodef{GPU}{Graphics Processing Unit}
\acrodef{CPU}{Central Processing Unit}
\acrodef{HB}{Hyperband}
\acrodef{ASHA}{Asynchronous Successive Halving Algorithm}
\acrodef{RASDA}{Resource-Adaptive Successive Doubling Algorithm}
\acrodef{BOHB}{Bayesian Optimization and HyperBand}
\acrodef{BO}{Bayesian optimization}
\acrodef{HPC}{High-Performance Computing}
\acrodef{CNN}{Convolutional Neural Network}
\acrodef{CFD}{Computational Fluid Dynamics}
\acrodef{PBT}{Population Based Training}
\acrodef{NCCL}{NVIDIA Collective Communications Library}
\acrodef{SGD}{Stochastic Gradient Descent}
\acrodef{ADAM}{Adaptive Moment Estimation}
\acrodef{GNS}{Gradient Noise Scale}
\acrodef{CV}{Computer Vision}
\acrodef{LLM}{Large Language Model}
\acrodef{MSE}{Mean-Squared Error}
\acrodef{AM}{Additive Manufacturing}
\acrodef{GB}{Gigabyte}
\acrodef{TB}{Terrabyte}
\acrodef{HEP}{High-Energy Physics}
\acrodef{EO}{Earth Observation}
\acrodef{EMA}{Exponential Moving Average}
\acrodef{AI}{Artificial Intelligence}
\acrodef{NLP}{Natural Language Processing}
\acrodef{DiT}{Diffusion Transformer}
\acrodef{CBS}{Critical Batch Size}
\acrodef{muP}{Maximal Update Parameterization}
\acrodef{GPT}{Generative Pre-Trained Transformer}
\acrodef{FID}{Fréchet Inception Distance}

\usepackage[final]{automl}
\usepackage[capitalize,noabbrev]{cleveref} %

\usepackage{natbib}
\title{OpenGNS: An Open Dataset of the Gradient Noise Scale across Vision, Language, and Diffusion Models}

\author[1]{\nameemail{Marcel Aach}{m.aach1@fz-juelich.de}}
\author[1]{\nameemail{Jiangtao Wang}{}}
\author[1]{\nameemail{Stefan Kesselheim}{}}
\author[1]{\nameemail{Andreas Lintermann}{}}

\affil[1]{Jülich Supercomputing Centre, Jülich, Germany}

\begin{document}

\maketitle

\begin{abstract}
The \ac{GNS} metric measures the signal-to-noise ratio of gradients in stochastic
gradient descent and, as shown by~\cite{mccandlish2018empirical}, can be used to
approximate the \ac{CBS} during \ac{NN} training. Despite its appearance
in technical reports of frontier models, no open dataset of \ac{GNS} measurements
across modern architectures exists, limiting systematic study of the metric. We
close this gap by releasing \textbf{OpenGNS}, a dataset of more than $400$ \ac{GNS} measurement trajectories
spanning three domains: Image classification with ResNets, autoregressive
language modeling with \acp{GPT}, and image synthesis with \acp{DiT}, covering
models from 11M to 1B parameters trained under \ac{muP} with dense batch-size and
learning-rate sweeps. Building on this dataset, we study how the \ac{GNS} depends on
hyperparameters, scale, and training progress, and how it relates to the \ac{CBS}.
We find that \ac{GNS} behaves as a consistent ``temperature of training'',
remaining stable across model sizes under \ac{muP} at equivalent loss, and
our results suggest task-dependent biases in its relation to the \ac{CBS}:
A tendency to overestimate the \ac{CBS} for classification and to
underestimate it for generative tasks, consistent in direction across common \ac{CBS} definitions. The dataset is available at \url{https://huggingface.co/datasets/olympique-marcel/OpenGNS} 
\end{abstract}

\acresetall

\section{Introduction}
The \acf{GNS}, introduced by \cite{mccandlish2018empirical}, provides a principled approach to understanding the trade-offs inherent in stochastic gradient descent. 
Intuitively, the \ac{GNS} quantifies the noise in the stochastic gradient: When gradient noise is high, larger batch sizes are beneficial, as averaging over more samples yields a more accurate approximation of the true gradient. When noise is already low, additional samples merely increase computational cost per update without improving gradient quality. This intuition connects the \ac{GNS} to the \acf{CBS}, which is the batch size that balances runtime and computational effort required to reach a given loss level, and is therefore highly relevant to the AutoML community.

Understanding and accurately estimating the \ac{CBS} has become increasingly important in the era of foundation models. Pre-training autoregressive \acp{LLM} and diffusion models from the \ac{CV} domain consumes substantial computational resources, often distributed across thousands of \acp{GPU}~\citep{singh2025openaigpt5card, grattafiori2024llama3herdmodels, nvidia2025cosmosworldfoundationmodel}. Making pre-training more efficient, therefore, reduces energy consumption and cost, and helps democratize \ac{AI} research for academic laboratories that lack industrial-scale compute. In the data-parallel training setting, the global batch size directly determines how many \acp{GPU} can be utilized in parallel, making batch-size selection a critical decision that affects training efficiency, resource utilization, and final model quality.

Despite its practical importance, the \ac{GNS} has received limited systematic investigation. While the original work by \cite{mccandlish2018empirical} predates the Transformer era, the metric does subsequently appear in technical reports of some frontier models, see, e.g., GPT-3~\citep{NEURIPS2020_1457c0d6}. Furthermore, open-source implementations exist for measuring the \ac{GNS} in both autoregressive and diffusion settings~\citep{crowson2024hourglass, gray2024normalization}. However, no comprehensive analysis has examined how the \ac{GNS} behaves across modern model architectures, training regimes, and hyperparameter configurations, and no open dataset of \ac{GNS} measurements is available to the community to study the metric and validate its predictions.

We close this gap with \textbf{OpenGNS}, an open dataset of \ac{GNS} measurements spanning three domains: image classification with ResNets on CIFAR-10, autoregressive language modeling with \acp{GPT} on FineWeb, and image synthesis with \acp{DiT} on ImageNet. The dataset covers eleven model configurations from 11M up to 1B parameters under \ac{muP}, trained over dense batch-size and learning-rate grids. Such a dataset is also directly relevant to the AutoML community. Full hyperparameter sweeps over batch size and learning rate are prohibitively expensive at the scale of modern foundation models, and methods that rely on cheap proxies (smaller models or shorter runs) depend on the assumption that low-fidelity behavior transfers to higher fidelities. Per-step trajectories of loss and \ac{GNS} across a dense grid of model sizes, batch sizes, and learning rates provide a quantitative basis for studying this assumption and for informing \ac{HPO} strategies for batch-size and learning-rate selection without re-running the underlying experiments. Alongside the dataset, we present an empirical study that demonstrates its utility and yields insights that can inform more efficient allocation of computational resources in large-scale training runs. 
In summary, we make the following three contributions: 
\begin{itemize}
    \item \textbf{The OpenGNS dataset:} We release \ac{GNS} measurements together with training and validation losses for more than 400 training runs spanning three domains, eleven model sizes, and dense batch-size/learning-rate grids under \ac{muP}. To our knowledge, this is the first open dataset of its kind.
    \item \textbf{GNS across hyperparameters:} The measurement of the \ac{GNS} depends strongly on the batch-size and learning-rate configuration. However, when both hyperparameters are jointly well-tuned, training runs exhibit consistent \ac{GNS} trajectories, validating the notion of the \ac{GNS} as a temperature of training~\citep{mccandlish2018empirical} also for modern Transformer architectures.
    \item \textbf{GNS across loss and model size:} Empirical tests examine how well the \ac{GNS} approximates the \ac{CBS} across different learning tasks, extending experiments from~\cite{mccandlish2018empirical} on CIFAR-10 to \ac{GPT}-style architectures and \acp{DiT}. The results suggest systematic differences depending on the task type: Overestimation for classification and underestimation for generative tasks. Scaling up model size under \ac{muP}, both metrics remain consistent across model sizes, enabling transfer of \ac{CBS} recommendations across widths at the same loss.

\end{itemize}

The paper is structured as follows: In Sec.~\ref{sec:background} we revisit the derivation of the \ac{GNS} metric and its measurement process. Section~\ref{sec:dataset} introduces the OpenGNS dataset, including the experimental setup, contents, and release format. Section~\ref{sec:empirical_study} presents the empirical findings enabled by the dataset. Related work is covered in Sec.~\ref{sec:related_work}. Finally conclusions are drawn in Sec.~\ref{sec:conclusion}.

\section{Background}\label{sec:background}
In this section, we revisit the theoretical derivation of the \ac{GNS} following~\cite{mccandlish2018empirical}, show how to measure it without overhead in data-parallel training settings, and define the \ac{CBS}. 

\subsection{Formal Derivation of the Gradient Noise Scale}\label{sec:gns_derivation}
The \ac{GNS} is based on a second-order Taylor expansion of the loss $L$ of a \ac{NN}. Let $\theta$ denote the \ac{NN} parameters, $G$ the true gradient, and $H$ the Hessian of the problem. Perturbing the parameters by a step of size $\epsilon$ along a direction $V$, the expansion of the loss around $\theta$ can be written as
\begin{equation}\label{eq:taylor_expansion}
    L(\theta - \epsilon V) \approx L(\theta) - \epsilon G^TV + \frac{1}{2} \epsilon^2 V^THV.
\end{equation}
Choosing $V = G$, the argument of the left-hand side of Eq.~\eqref{eq:taylor_expansion} becomes $\theta - \epsilon G$, i.e., the update rule of gradient descent with learning rate $\epsilon$. With access to the true gradient and Hessian, it is possible to compute the optimal step size that minimizes the loss. \ac{SGD}, however, only provides access to a gradient estimate $\mathbb{E}[G_{est}]$ with 
\begin{align}
    \mathbb{E}[G_{est} (\theta)] = G(\theta). 
\end{align}
The covariance matrix of $G_{est}$ is defined as
\begin{align}
    cov(G_{est} (\theta)) = \frac{1}{B} \Sigma (\theta),
\end{align}
where $B$ is the batch size and $\Sigma$ is the per-example covariance matrix. Computing the expected loss based on this, Eq.~\eqref{eq:taylor_expansion} can then be rewritten as
\begin{eqnarray}
    \mathbb{E}[L(\theta - \epsilon G_{est})] &\approx& L(\theta) - \epsilon |G|^2 +\nonumber\\ 
    &&\frac{1}{2} \epsilon^2 \left( G^T HG + \frac{tr(H\Sigma)}{B} \right).
\end{eqnarray}
It can be shown that in this case the optimal batch size, under assumption of the optimal step size, is attained when the batch size is equal to the \ac{GNS}, i.e., 
\begin{equation}\label{eq:exact_noise_scale}
    B_{noise} = \frac{tr(H\Sigma)}{G^THG}.
\end{equation}
Under the further assumption that the optimization is perfectly conditioned, the Hessian becomes a multiple of identity matrix and Eq.~\eqref{eq:exact_noise_scale} reduces to 
\begin{equation}\label{eq:simple_noise_scale}
    B_{simple} = \frac{tr(\Sigma)}{G^TG},
\end{equation}
which is easier to measure during the training, as computing the Hessian is usually expensive.  The quantity $B_{simple}$ then essentially becomes a measure of noise-to-signal of the gradient. When we measure the \ac{GNS} empirically throughout this work, we make use of the $B_{simple}$ approximation and refer to is as \ac{GNS}.

\subsection{Measuring the Gradient Noise Scale}\label{sec:measuring_gns}
Multiple options for measuring the simplified version of the \ac{GNS} in Eq.~\eqref{eq:simple_noise_scale} exist. An overview of the different techniques is given in~\cite{gray2024normalization}. In this paper, the method presented by~\cite{mccandlish2018empirical} for data-parallel training environments is used. It is based on the idea that, given gradients at a small and large batch size, the expected norms can be computed as
\begin{align}\label{eq:norms_data_parallel}
    \mathbb{E}[|G_{small}|^2] &= |G|^2 + \frac{1}{B_{small}}tr(\Sigma), \\
    \mathbb{E}[|G_{big}|^2] &= |G|^2 + \frac{1}{B_{big}}tr(\Sigma). \label{eq:norms_data_parallel_2}
\end{align}
Solving for $|G|^2$ and $tr(\Sigma)$ by substituting terms from Eq.~\eqref{eq:norms_data_parallel_2} into Eq.~\eqref{eq:norms_data_parallel} then yields
\begin{eqnarray}
    |G|^2 &\approx& \frac{1}{B_{big} - B_{small}}\cdot \nonumber\\
    &&\left( B_{big}|G_{B_{big}}|^2 - B_{small}|G_{B_{small}}|^2 \right) \\
    tr(\Sigma) &\approx& \frac{1}{1 \ / B_{small} - 1 \ / B_{big}}\cdot\nonumber\\
    &&\left( |G_{B_{small}}|^2 - |G_{B_{big}}|^2 \right).
\end{eqnarray}
Recall that in data-parallel training, the global gradient update is typically computed as 
\begin{equation}\label{eq:grad_avg}
    G_{global} (\theta) = \frac{1}{N} \sum_{i=1}^N G_i (\theta),
\end{equation}
where $N$ is the number of devices (\acp{GPU}) and $G_i$ is the local gradient update on each \ac{GPU}. Therefore, by setting $G_{big}=G_{global}$, $G_{small}=G_{i}$, and $B_{big} = N \cdot B_{small}$, the necessary measurements for computing $B_{simple}$ are available with no additional overhead. To account for the nosiness of these measurements, both values are measured every few training steps and separate \acp{EMA} with a decay parameter $\alpha$ are fitted to stabilize them, see Sec.~\ref{sec:gns_measurement_params} for an analysis of the impact on the measurements. Other options for measuring the \ac{GNS} in Transformer models include the methods proposed by \cite{gray2024normalization} which measure gradient norms on a per-example basis, but require custom \ac{GPU} kernels. 

\subsection{Definition of the Critical Batch Size}\label{sec:crit_bs_definition}
In the literature, the term \ac{CBS} is often used with different meanings. In the ideal linear scaling case, an increase of the batch size by a factor $k$ results in a decrease by of optimization steps, required to reach a certain loss threshold, by the same factor. The "original" definition by ~\cite{mccandlish2018empirical} defines the \ac{CBS} as the batch size, where the scaling efficiency drops below $50\%$. In ~\cite{zhang2024how}, the authors tighten this bound and define the \ac{CBS} as the batch size, where an overhead of only $20\%$ is allowed. Different from the \ac{CBS} is the ``Optimal Batch Size'', which is sometimes defined as the smallest batch size that can be scaled to without any overhead~\citep{zhang2024how}, or the batch size that achieves the lowest validation loss under a given data horizon~\citep{filatov2025timetransferoptimallearning}. Another value is the ``Maximum Stable Batch Size'', which is the largest batch size that does not break the training, i.e., the batch size that can recover a certain loss level irresistible of the number of optimization steps/overhead. In general Optimal Batch Size $<$ \acf{CBS} $<$ Maximum Stable Batch Size.  

\section{OpenGNS Dataset}\label{sec:dataset}
This section describes the OpenGNS dataset. Section~\ref{sec:experimental_setup} introduces the three model architectures and experimental setup that produced the measurements, and Sec.~\ref{sec:dataset_contents} details the contents and release format of the data.

\subsection{Experimental Setup}\label{sec:experimental_setup}

To investigate the \ac{GNS}, we experiment with a supervised learning approach from the image classification domain and two self-supervised learning approaches, one each from the \ac{NLP} and \ac{CV} domains. Specifically, we use a ResNet18~\citep{he2016} for image classification, an autoregressive language model for \ac{NLP}, and a \ac{DiT} image model for \ac{CV}. All models are trained with the AdamW optimizer~\citep{loshchilov2018decoupled}. The full hyperparameter grid is summarized in Tab.~\ref{tab:hyperparameters}.

The image classification workload trains on the CIFAR-10 dataset~\citep{cifar_10} for 200 epochs. Following the original \ac{muP} implementation~\citep{tensor_programm_v}, the channel widths are scaled by factors of $1\times$, $2\times$, and $4\times$ relative to the base architecture. We use a cosine learning rate decay schedule that performs one decay step per epoch. For a final generalization test, we leverage the CIFAR-10.1 dataset\footnote{CIFAR-10.1: \url{https://github.com/modestyachts/CIFAR-10.1}}.
For the autoregressive Transformer model from the \ac{NLP} domain, we use the nano-muP implementation~\citep{cerebras2024mupguide}, which is based on NanoGPT~\citep{Karpathy2022}. The models are trained for $20$B tokens on the FineWeb dataset ($100$B token version)~\citep{penedo2024finewebdatasetsdecantingweb}, which roughly matches the compute-optimal training duration for the largest model configuration according to the Chinchilla scaling law~\citep{hoffmann2022an}. 
The diffusion models are based on \acp{DiT}~\citep{Peebles2022DiT} and use the \ac{muP} implementation of~\citep{zheng2025scalingdiffusiontransformersefficiently}. Models are trained on ImageNet~\citep{imagenet} for $51$M images, corresponding to $\approx 800{,}000$ steps at batch size $2^6$. The validation loss is computed on the COCO validation set, following the practice in~\citep{stable_diffusion_3}. 

All models are trained on the Leonardo Booster \ac{HPC} system featuring four NVIDIA A100 \acp{GPU} per node, connected via InfiniBand. The \ac{GNS} is measured by adding a custom hook to the PyTorch-DDP framework, based on the implementations of~\citep{crowson2024hourglass} and~\citep{gray2024normalization}. The \ac{GPT} and \ac{DiT} models are trained using two nodes (8 \acp{GPU}), while the ResNet models use a single node (4 \acp{GPU}). In total, $\approx 200{,}000$ \acp{GPU} hours were used for development and production runs. 

\begin{table}[ht]
\centering
\caption{Hyperparameter grids for the three experimental workloads. There are 180 configurations for the ResNet model, 178 for the \ac{GPT} model and 75 for the \ac{DiT} models, for a total of $433$ training runs.}
\label{tab:hyperparameters}
\begin{tabular}{clccc}
\toprule
\textbf{Hyperparameter} & & \textbf{ResNet18} & \textbf{\ac{GPT}} & \textbf{DiT} \\
\midrule
Dataset & & CIFAR-10 & FineWeb & ImageNet \\
Training data & & $10$M images & $20$B tokens & $51$M images \\
Validation data & & 10{,}000 images & $100$M tokens & $40{,}504$ images \\
\midrule
Widths & & $[1, 2, 4]$ & $[256, 512, 1{,}024, 2{,}048, 2{,}560]$ & $[144, 288, 576]$ \\
Depth & & 18 & 12 & 28  \\
Attn. head size & & - & 128 & 72  \\
Parameters & & 11M, 45M, 179M & 22M, 64M, 203M, 707M, 1B & 11M, 43M, 169M \\
\midrule
Batch size & & $2^5$--$2^{14}$ images & $2^{15}$--$2^{21}$ tokens & $2^6$--$2^{10}$ images \\
Learning rate & & $10^{-5}$--$10^{-1}$ & $2^{-11}$--$2^{-6}$ & $2^{-13}$--$2^{-9}$ \\
Optimizer & & AdamW  & AdamW & AdamW \\
$\beta_1$ & & 0.9 & 0.9 & 0.9\\
$\beta_2$ & & 0.99 & 0.95 & 0.99\\
Weight decay & & 0.0 & 0.01 & 0.01  \\
LR schedule & & Cosine (epoch based) & Cosine (step based) & Constant \\
Final LR & & $10\%$ & $10\%$ & - \\
LR warmup & & None & $5\%$ of training duration & None \\
$\alpha$ for \ac{GNS} & & 0.995 & 0.95 & 0.995 \\
$B_{\text{big}}/B_{\text{small}}$ & & 4 & 8 & 8 \\
\bottomrule
\end{tabular}
\end{table}

\subsection{Dataset Contents}\label{sec:dataset_contents}

Each (model, batch size, learning rate) combination is trained with a single random seed. For the large \ac{GPT} models (widths $2{,}048$ and $2{,}560$), the small batch sizes $2^{15}$, $2^{16}$, and $2^{17}$ are skipped due to long training times. For \acp{DiT}, a small number of combinations ($9$ out of $75$) exhibit training instabilities after roughly half the training time and diverge to NaN values, particularly at larger learning rates, in line with what is reported by the original \ac{DiT}-\ac{muP} implementation\footnote{DiT-muP implementation: \url{https://github.com/ML-GSAI/Scaling-Diffusion-Transformers-muP}}. Similarly, $12$ out of $210$ \ac{NLP} runs diverge through loss spikes, all at the two largest learning rates. We nevertheless include these runs in the dataset, recorded up to the point of divergence, as the onset of instability itself marks a stability boundary of the batch-size--learning-rate grid.

For each training run, we record the training and validation loss together with three \ac{GNS}-related quantities at regular measurement intervals: The simplified noise scale itself, $B_\text{simple}$ (stored as \texttt{gns}), the squared gradient norm $|G|^2$ (\texttt{gns\_norm}), and the trace of the per-example gradient covariance $\text{tr}(\Sigma)$ (\texttt{gns\_var}), see Fig.~\ref{fig:gns_lr_dependency}. Releasing the two raw components separately allows users to study them separately and reconstruct the \ac{GNS} themselves. 

The dataset is hosted on HuggingFace\footnote{OpenGNS dataset: \url{https://huggingface.co/datasets/olympique-marcel/OpenGNS/tree/main}} and split into one Parquet file per workload: \texttt{cv.parquet}, \texttt{nlp.parquet}, and \texttt{diffusion.parquet}. All three files share the columns \texttt{batch\_size}, \texttt{width} or equivalent model-size identifier, an iteration counter, \texttt{samples\_seen}, and the five measurement columns above. \texttt{nlp.parquet} additionally stores both the \texttt{peak\_lr} hyperparameter and the instantaneous \texttt{lr}, which varies due to the cosine decay schedule. \texttt{cv.parquet} adds validation and test accuracy. \texttt{diffusion.parquet} stores the learning rate on a $\log_2$ scale and includes a second validation loss on COCO images (\texttt{coco\_val/loss}).

\section{Empirical Study of the GNS}\label{sec:empirical_study}
Using the OpenGNS dataset from Sec.~\ref{sec:dataset}, we empirically study the \ac{GNS}. Section~\ref{sec:gns_over_hyperparam} analyzes its dependence on batch size and learning rate, while Sec.~\ref{sec:gns_cbs_loss} examines its relationship to the \ac{CBS} across training loss and model size under \ac{muP}.

\subsection{Gradient Noise Scale over Hyperparameters}\label{sec:gns_over_hyperparam}
\begin{figure}
    \centering
    \includegraphics[width=0.32\linewidth]{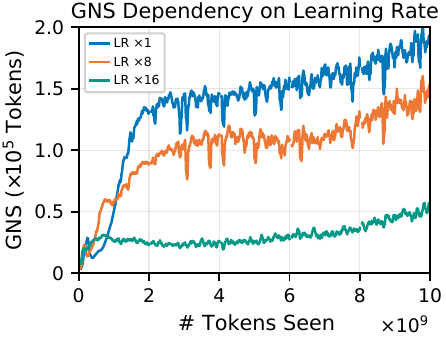}
    \includegraphics[width=0.64\linewidth]{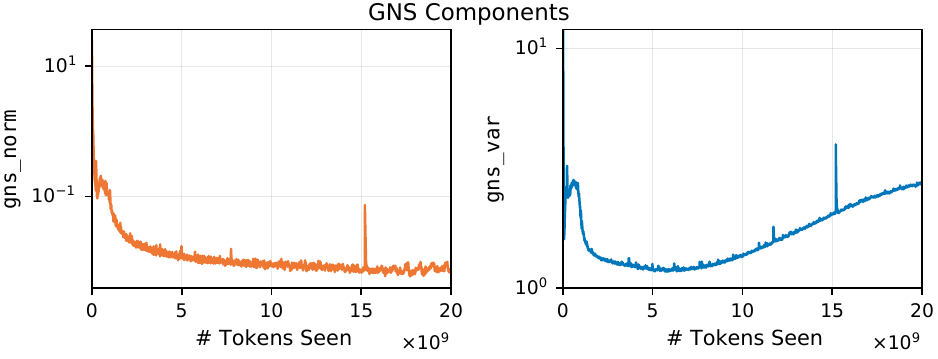}
    \caption{Left: Dependency of the \ac{GNS} on the learning-rate hyperparameter, for a fixed batch-size, model-size, and training task (\ac{GPT} model on FineWeb data). Linear axis scaling is employed for better visibility. Middle and Right: The two raw components of the \ac{GNS} (the squared gradient norm and the trace of the per-example gradient covariance) over the course of training for a representative \ac{GPT} model run (Width=512, BS=$2^{20}$, LR= $2^{-7}$).}\label{fig:gns_lr_dependency}
\end{figure}

For the same training objective, \ac{GNS} measurements can vary substantially depending on the choice of hyperparameters. As explained in the derivation of the \ac{GNS} in Sec.~\ref{sec:gns_derivation}, the measurement is sensitive to the choice of learning rate relative to the batch size. Figure~\ref{fig:gns_lr_dependency} depicts this dependence, as for the same model architecture and training task (\ac{GPT} model on FineWeb data), the measured \ac{GNS} varies by nearly an order of magnitude when the learning rate is scaled by factors of 8 and 16 from the baseline.

As \cite{mccandlish2018empirical} point out, the \ac{GNS} depends on the learning rate $\epsilon$ and the batch size $B$ through what they term the training temperature. This reflects the idea of the loss function of an \ac{NN} as a potential energy function. When the learning rate is too small relative to the batch size, the optimizer explores only a narrow region around its current position, leading to artificially inflated \ac{GNS} measurements. However, when the learning rate is appropriately scaled with the batch size, configurations that reach similar validation losses settle into the same temperature regime and produce comparable \ac{GNS} trajectories.

\begin{figure*}
    \includegraphics[width=0.33\linewidth]{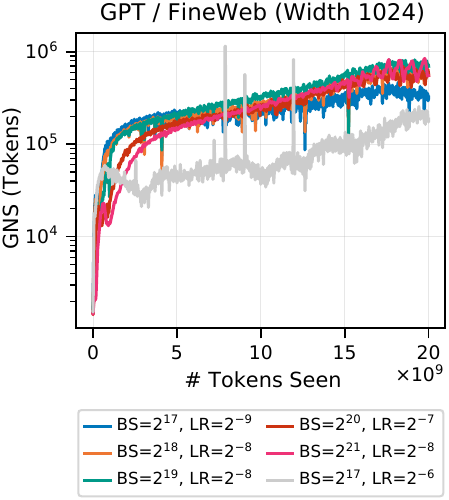}
    \includegraphics[width=0.33\linewidth]{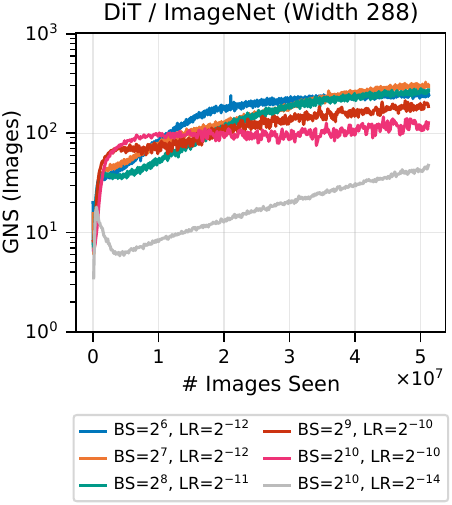}
    \includegraphics[width=0.33\linewidth]{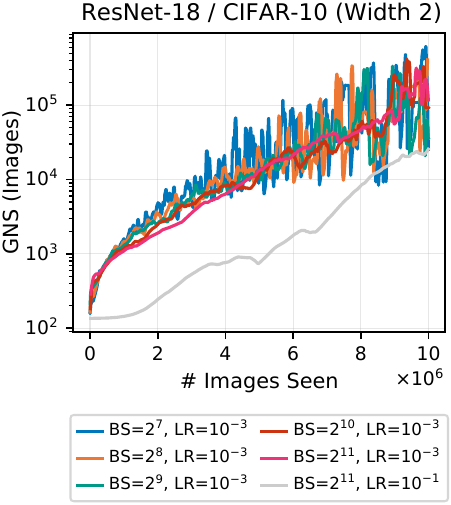}
    \caption{\ac{GNS} as a the temperature of training. Over the course of the training run, batch-size and learning-rate combinations that reach similar final validation losses share similar temperature trajectories. Note that for the \ac{GPT} and \ac{DiT} it is possible to recover the Adam-specific scaling rules (learning rate $\times 2$, batch size $\times 4$, blue and green lines). A peak in the learning rate that occurs roughly at the final \ac{GNS} level, as also predicted by~\cite{li2024surge}, can be identified. The \ac{GNS} curves for smaller batch sizes employ additional smoothing for better visibility.}\label{fig:gns_as_temperature}
\end{figure*}

The OpenGNS dataset enables us to systematically investigate this dependence by comparing \ac{GNS} trajectories across a wide variety of batch-size and learning-rate combinations for all three training tasks. We choose for each batch size the learning rate that achieves the lowest validation loss and define this as the optimal configuration. As shown in Fig.~\ref{fig:gns_as_temperature}, the \ac{GNS} trajectory throughout the training depends significantly on the specific hyperparameter configuration chosen. For each task (autoregressive language modeling with a \ac{GPT} on FineWeb data, image classification with a ResNet-18 on CIFAR-10, and diffusion modeling with a \ac{DiT} on ImageNet) we observe that configurations reaching worse final validation losses (highlighted in gray) trace \ac{GNS} trajectories that deviate substantially from those reaching the lowest validation losses.

However, as the colored curves of Fig.~\ref{fig:gns_as_temperature} emphasize, well-optimized training runs share consistent \ac{GNS} trajectories, which represent configurations that achieve comparably low final validation losses. Despite varying the batch size by up to a factor of 16 (from $2^{17}$ to $2^{21}$ tokens for \ac{GPT}, from $2^{6}$ to $2^{10}$ images for \ac{DiT}, and from $2^{7}$ to $2^{11}$ images for ResNet), the \ac{GNS} trajectories align when each configuration is paired with an appropriate learning rate.

For some hyperparameter combinations, we are also able to recover the approximate optimal learning-rate--batch-size scaling rules. In particular, the \ac{GPT} and \ac{DiT} experiments, see Fig.~\ref{fig:gns_as_temperature}, show cases where scaling the batch size by a factor of four while doubling the learning rate (blue and green lines) produces nearly identical \ac{GNS} trajectories. This is consistent with the Adam-specific scaling behavior documented in prior work, where the optimal learning rate scales as $\epsilon \propto \sqrt{B}$, see~\citep{hilton2022batch, malladi2022on}. Additionally, in line with the predictions of \cite{li2024surge}, we observe that the learning-rate peak in batch-size--learning-rate scaling for well-optimized \ac{GPT} runs occurs at batch sizes ($2^{19}$, green line; $2^{20}$, red line) close to the final \ac{GNS} values, between $0.5\cdot 10^6$ and $10^6$ tokens.

\subsection{Gradient Noise Scale and Critical Batch Size over Loss and Model Size}\label{sec:gns_cbs_loss}
We next investigate how well the \ac{GNS} approximates the \ac{CBS} and how both metrics depend on the model size when scaling width under \ac{muP}.

The \ac{CBS} is calculated by examining different loss levels and quantifying the required overhead in optimization steps when, compared to a base batch size, the batch size to reach the same loss level is doubled. Following the definition of~\cite{zhang2024how}, we define the \ac{CBS} as the batch size, where the scaling efficiency overhead exceeds $20\%$, see App.~\ref{sec:cbs_computation} for a more detailed explanation and App.~\ref{sec:cbs_sensitivity} for a sensitivity analysis with respect to this threshold choice.

\begin{figure}
    \centering
    \includegraphics[width=0.32\linewidth]{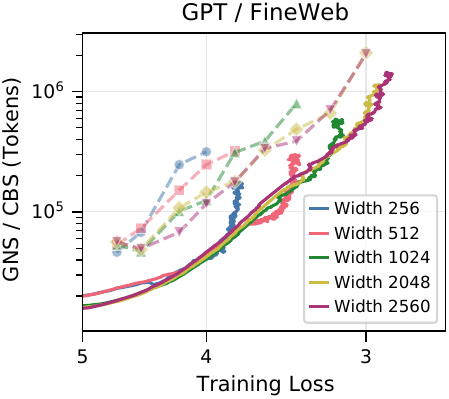}
    \includegraphics[width=0.32\linewidth]{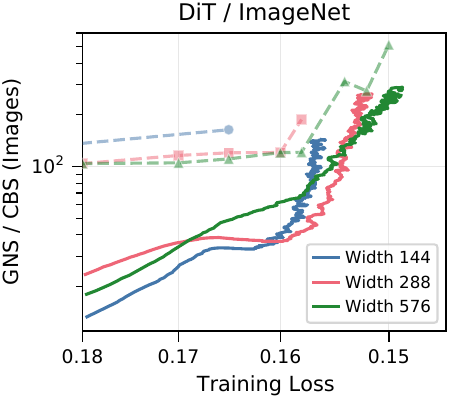}
    \includegraphics[width=0.32\linewidth]{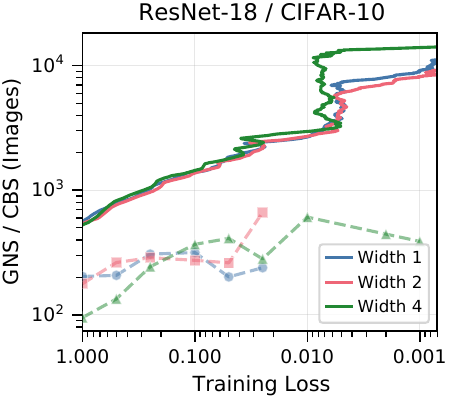}
    \caption{The \ac{GNS} (solid lines) and \ac{CBS} (dashed, light lines) as a function of training loss across model sizes for well-optimized batch-size and learning-rate combinations. The \ac{GNS} remains roughly consistent across model sizes at the same training loss, suggesting that measurements from small models can inform batch-size selection for large models. Large models achieve higher \ac{GNS} values primarily because they reach lower training losses than small models. 
    While the \ac{GNS} tracks the overall trend of the \ac{CBS}, it tends to underestimate the \ac{CBS} for generative tasks (\ac{GPT}, \ac{DiT}) and overestimate it for the classification task (ResNet).}
    \label{fig:gns_cbs_loss}
\end{figure}

Figure~\ref{fig:gns_cbs_loss} shows the \ac{GNS} and \ac{CBS} as a function of the training loss across different model sizes for all three tasks. The \ac{GPT} models exhibit the highest \ac{GNS} and \ac{CBS} values, reaching up to $10^6$ tokens, followed by the image classification ResNets with values up to tens of thousands of images. The diffusion models show comparatively smaller values, on the order of hundreds of images.

A key observation is that the \ac{GNS} values remain roughly consistent across model sizes at the same training loss level. This is consistent with the theoretical expectation from~\cite{mccandlish2018empirical}, i.e., the noise scale should not exhibit a strong dependence on the model size directly, but rather primarily on the achieved loss. Large models attain higher \ac{GNS} values mainly because they reach lower training losses than smaller models. This consistency across model sizes under \ac{muP} suggests that \ac{GNS} measurements from smaller proxy models could inform both learning-rate and batch-size selection for larger models, potentially reducing the cost of hyperparameter optimization for large-scale training runs.

While the \ac{GNS} tracks the overall trend of the \ac{CBS} across all tasks, we observe systematic differences in their relationship. For the generative tasks (\ac{GPT} and \ac{DiT}), the \ac{GNS} tends to underestimate the \ac{CBS}, consistent with recent findings in~\cite{merrill2025critical}. At the same time, for the classification task (ResNet on CIFAR-10), the \ac{GNS} tends to overestimate the \ac{CBS}, aligning with the original measurements reported in~\cite{mccandlish2018empirical}. One reason for these systematic discrepancies could be the simplified assumption of a well-conditioned Hessian used to derive $B_{\text{simple}}$ in Eq.~\eqref{eq:exact_noise_scale}, which may hold to different degrees across task types. For the \ac{NLP} task in particular, the heavy-tailed class imbalance of the token distribution is known to induce exactly this kind of ill-conditioning~\cite{kunstner2024heavytailed}. The direction of the \ac{GNS}-\ac{CBS} deviations persists under alternative overhead thresholds of $10\%$, $30\%$, and $50\%$, although their magnitude varies with the threshold and the gap for the classification task largely closes under the original $50\%$ definition of~\cite{mccandlish2018empirical}, see App.~\ref{sec:cbs_sensitivity}.

\section{Related Work on Gradient Noise Scale}\label{sec:related_work}
Apart from~\cite{mccandlish2018empirical}, an earlier definition of a noise scale of the gradient was proposed by~\cite{smith2018}, modeling the training process of an \ac{NN} as a stochastic differential equation. Frameworks like Pollux~\citep{pollux} and KungFu~\citep{kungfu} feature the capability to use the \ac{GNS} metric for selecting a suitable batch size during the training, however, they do not perform a systematic study of its dependence on hyperparameters or model size, nor do they release the underlying measurements. In~\cite{gray2023efficient, gray2024normalization}, methods to compute the \ac{GNS} analytically in both approximate and exact forms are introduced, with a focus on \acp{LLM}. The authors of~\cite{crowson2024hourglass} use the \ac{GNS} for scheduling the batch size during the final $10\%$ of training of a \ac{DiT}, increasing the batch size up to a factor of $10$.

More recently, measurements of the \ac{GNS} for long training horizons (up to $1$T tokens) have been performed in~\cite{merrill2025critical} and~\cite{li2025efficienthyperparametertuningtrajectory}, making use of pretrained model checkpoints to compute the \ac{GNS} at different intervals of training. Coupled batch-size and learning-rate schedules were recently introduced in~\cite{meterez2025seesawacceleratingtrainingbalancing}, but work only below the \ac{CBS} and thus require knowing its value a priori. Common to all of these works is that, while they report \ac{GNS} measurements for individual training runs, none release the underlying trajectories in a form that allows other researchers to reuse them for follow-up study. OpenGNS differs in this respect: It provides per-step \ac{GNS} measurements together with loss curves across a dense grid of batch sizes, learning rates, and model sizes, spanning image classification, autoregressive language modeling, and diffusion-based image synthesis. Open releases of intermediate training checkpoints~\citep{pythia, groeneveld-etal-2024-olmo, nezhurina2025opensciref001openreproduciblereference} provide rich training-trajectory data, but focus exclusively on \ac{NLP} models and do not include \ac{GNS} measurements. It should be noted that~\cite{merrill2025critical} found that for \acp{LLM} at $1$B and $7$B parameter scale, the \ac{GNS} underestimates the \ac{CBS} by multiple orders of magnitude. This underestimation is consistent with our results for generative tasks, though not in this order of magnitude. Their \ac{CBS} definition also uses a loss-recovery criterion rather than accounting for a scaling-efficiency overhead as proposed by~\cite{mccandlish2018empirical}, which intuitively aligns more with the ``Maximum Stable Batch Size'', see Sec.~\ref{sec:crit_bs_definition}.

\section{Summary and Conclusion}\label{sec:conclusion}

This study contributes \textbf{OpenGNS}, an open dataset of \ac{GNS} measurements across three domains: image classification, autoregressive language modeling, and diffusion-based image synthesis. The empirical study enabled by this dataset confirms prior observations regarding \ac{CBS} differences across tasks: \acp{GPT} benefit from batch sizes in the millions of tokens, while \acp{DiT}, despite often reaching billions of parameters in state-of-the-art systems, operate efficiently at much smaller batch sizes on the order of hundreds of images (corresponding to roughly tens of thousands of patch tokens).

Our analysis validates the finding of~\cite{mccandlish2018empirical} that the \ac{GNS} behaves as a temperature of training, remaining consistent across runs that reach similar validation losses. This work extends this observation to modern Transformer and diffusion architectures and demonstrates that, combined with \ac{muP}, the consistency holds across model sizes. We find that the relationship between \ac{GNS} and \ac{CBS} is not always as straightforward as the theory might suggest: In our experiments, the \ac{GNS} appears to act more as a lower bound for the \ac{CBS} in the generative tasks and more as an upper bound in the classification task. Despite certain limitations in the theoretical assumptions and experimental scale (see App. ~\ref{sec:limitations}), the cross-size consistency of the \ac{GNS} under \ac{muP} suggests that measurements from smaller proxy models could inform batch-size selection for larger ones, reducing the computational cost of hyperparameter optimization at scale, which is relevant for the AutoML community. Releasing the raw gradient norm and variance measurements alongside the noise scale further enables deeper investigation of training dynamics across architectures and training regimes.
The increase of the \ac{GNS} as training progresses also points to a practical avenue for resource allocation: Since smaller global batch sizes correspond to fewer \acp{GPU} required for data-parallel training, \ac{GNS}-guided batch-size scheduling could allow for more efficient resource utilization, using a low number of \acp{GPU} in the early phases of training when small batches are sufficient, and scaling up to larger batches and more devices as training progresses. By releasing OpenGNS, we hope to encourage further investigation of such strategies and a deeper theoretical understanding of the \ac{GNS} metric.

\section*{Broader Impact Statement}

This paper presents work whose goal is to advance the field of Machine Learning. Although our evaluations require significant computational resources, the insights we provide in this work will enable more efficient training practices in the future, ultimately reducing resource waste and energy consumption.

\begin{acknowledgements}
We acknowledge the EuroHPC Joint Undertaking for awarding this project access to the EuroHPC supercomputer LEONARDO, hosted by CINECA (Italy) and the LEONARDO consortium through an EuroHPC Regular Access call. The research leading to these results is funded by the German Federal Ministry for Economic Affairs and Energy within the project “NXT GEN AI METHODS – Generative Methoden für Perzeption, Prädiktion und Planung". The authors would like to thank the consortium for the successful cooperation. 
\end{acknowledgements}

\bibliography{example_paper}

\newpage

\appendix
\onecolumn

\section{Limitations}\label{sec:limitations} 
The theoretical framework of the \ac{GNS} relies on several simplifying assumptions. It is derived using a second-order Taylor expansion and assumes a well-conditioned Hessian, see Eq.~\eqref{eq:exact_noise_scale}. While this approximation proves empirically useful, the systematic differences observed between the \ac{GNS} and the \ac{CBS} across task types (overestimation for classification, underestimation for generative tasks) could be a sign of ill-conditioning. Second, all experiments use the AdamW optimizer rather than plain \ac{SGD}, for which the original theory was derived. AdamW has largely replaced \ac{SGD} as the most popular optimizer. However, experiments in ~\citep{mccandlish2018empirical}, show that the \ac{GNS} values remain consistent across loss levels for both \ac{SGD} and Adam, suggesting the metric also works for adaptive gradient methods. Some works also make use of a preconditioned or non-Euclidean version of the \ac{GNS} for this reason~\citep{pollux, li2025efficienthyperparametertuningtrajectory, naganuma2026adaptivebatchsizesusing}. Note that our measurements are taken on the raw gradients before AdamW applies momentum and preconditioning, and the optimizer state is not logged during training. The released data therefore does not capture optimizer-state effects, and preconditioned noise-scale variants cannot be reconstructed from it, which we leave to future work. Third, the \ac{GNS} is defined purely in terms of the training optimization dynamics and provides no direct information about a generalization. 
Finally, while the experiments involve an extensive grid search using substantial computational resources to thoroughly explore different hyperparameter and model size configurations, the largest models reach only one billion parameters trained on tens of billions of tokens. State-of-the-art generative models often contain $>100$B parameters and are trained on trillions of tokens, rendering experiments at such scales infeasible. Nevertheless, the consistency of \ac{GNS} trajectories across model sizes under \ac{muP} suggests that the findings could be extrapolated to even larger scales.

\section{Gradient Noise Scale Measurement Parameters}
\label{sec:gns_measurement_params}

As described in Sec.~\ref{sec:measuring_gns}, the measurement of the \ac{GNS} using the data-parallel method depends on two parameters: The ratio of $B_{\text{big}}$ and $B_{\text{small}}$ as well as the \ac{EMA} decay parameter $\alpha$. We first investigate how varying these parameters affects the \ac{GNS} measurements.

\begin{figure}[ht]
    \centering
    \includegraphics[width=0.7\linewidth]{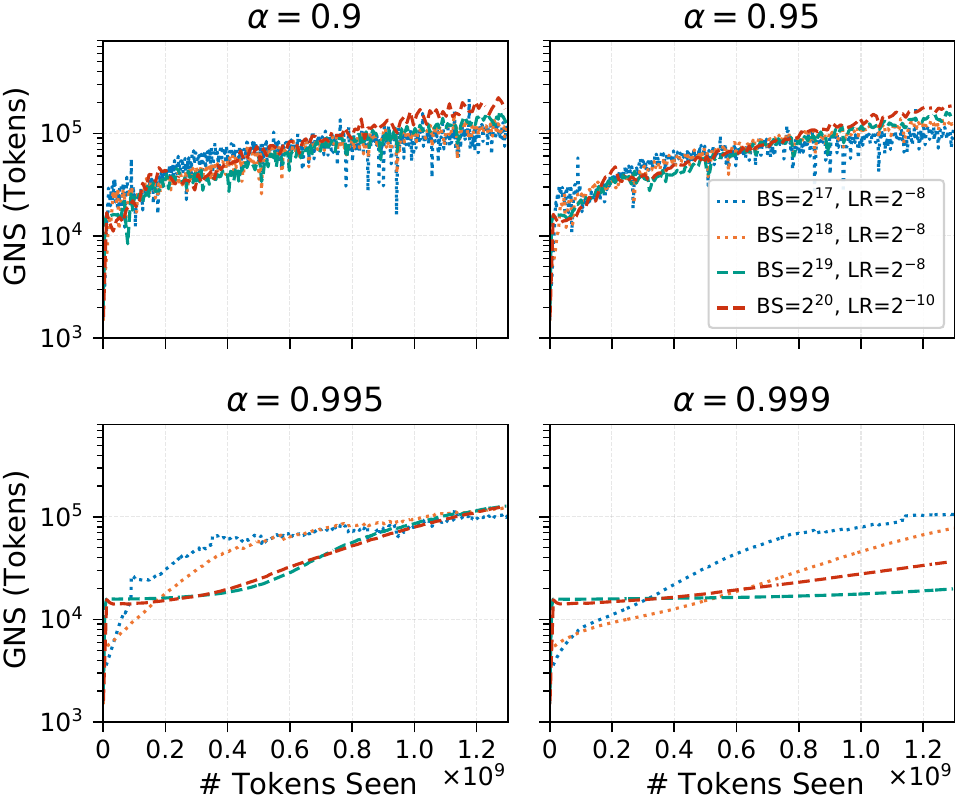}
    \caption{Influence of the \ac{EMA} decay parameter $\alpha$ on \ac{GNS} measurements (\ac{GPT} model trained on FineWeb data). Small values ($\alpha = 0.9$) yield noisy estimates, while values close to one ($\alpha = 0.999$) smooth excessively.}
    \label{fig:ema_values}
\end{figure}
\begin{figure}[ht!]
    \centering
    \includegraphics[width=0.5\linewidth]{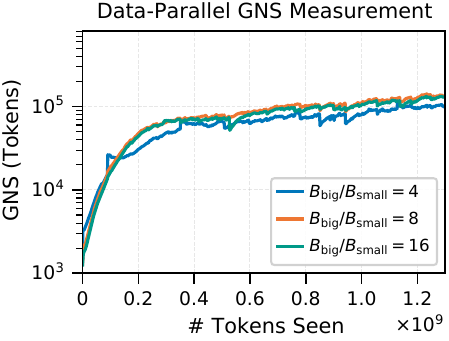}
    \caption{Influence of the batch size ratio $B_{\text{big}}/B_{\text{small}}$ on \ac{GNS} measurements with the data-parallel method of~\citep{mccandlish2018empirical} for a \ac{GPT} model trained on FineWeb data. The estimates remain consistent across different ratios, indicating robustness to this parameter choice.}
    \label{fig:bs_gns_relation}
\end{figure}

As expected, low \ac{EMA} values, e.g., $\alpha = 0.9$, result in noisier approximations that closely track rapid changes, while high values, e.g., $\alpha = 0.999$, produce smoother estimates but respond slower to variations in the \ac{GNS}, see Fig.~\ref{fig:ema_values}. Based on these observations, we set $\alpha$ between $0.95$ and $0.995$ depending on the task. Regarding the values of $B_{\text{big}}$ and $B_{\text{small}}$, \citep{gray2024normalization} find that small $B_{\text{small}}$ values reduce the standard error. As shown in our experiments in Fig.~\ref{fig:bs_gns_relation}, the ratio $B_{\text{big}}/B_{\text{small}}$ does not significantly affect the measured \ac{GNS} values for a Transformer model. This robustness simplifies practical deployment as the measurement can be performed reliably across different data-parallel configurations without requiring careful tuning of the batch size ratio. Our experiments use ratios of $4$ to $8$.

\section{Computing the Critical Batch Size}\label{sec:cbs_computation}

We compute the \ac{CBS} as follows. For each run we smooth the training loss with an exponentially weighted moving average whose span scales as $1/\sqrt{B}$, so that the smoothing window is roughly constant in samples seen. For a given target loss $L^\star$, we record the number of samples $N(w, B, \epsilon;\, L^\star)$ each run needs to first reach $L^\star$, and for every batch size we keep only the best-performing learning rate, $N^\star(w, B;\, L^\star) = \min_\epsilon N(w, B, \epsilon;\, L^\star)$. Using the smallest swept batch size $B_{\min}$ as reference, the scaling efficiency is $S(B) = N^\star(B_{\min}) / N^\star(B)$. Following~\citep{zhang2024how}, the \ac{CBS} is the batch size at which $S$ first drops below $1/1.2$, i.e.\ a $20\%$ scaling efficiency overhead, obtained by log-linear interpolation between the two adjacent batch sizes on our grid.

\section{Sensitivity to Different Critical Batch Size Thresholds}\label{sec:cbs_sensitivity}
\begin{figure}
    \centering
    \includegraphics[width=0.32\linewidth]{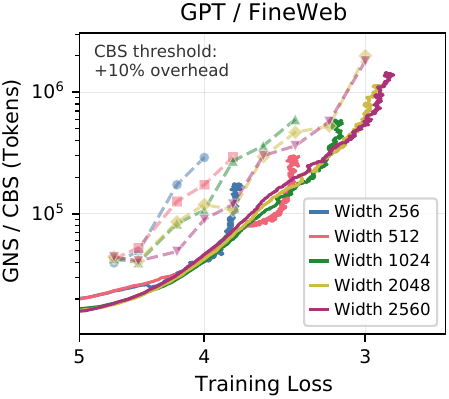}
    \includegraphics[width=0.32\linewidth]{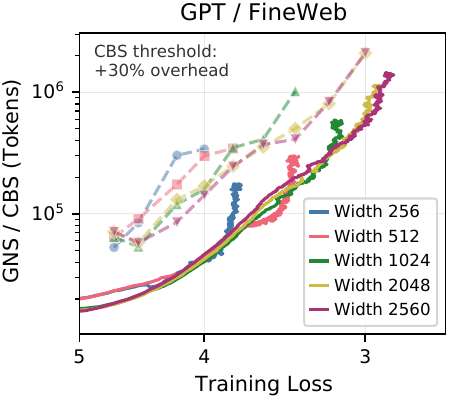}
    \includegraphics[width=0.32\linewidth]{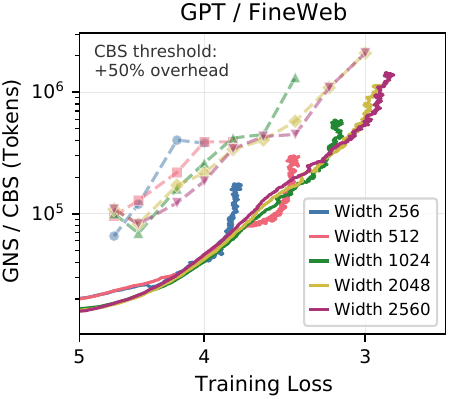}
    \includegraphics[width=0.32\linewidth]{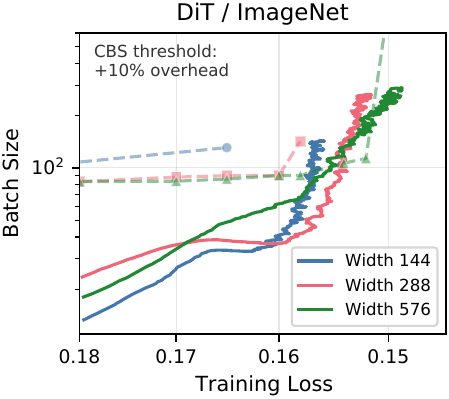}
    \includegraphics[width=0.32\linewidth]{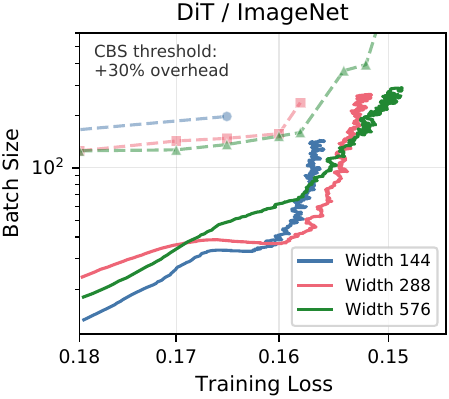}
    \includegraphics[width=0.32\linewidth]{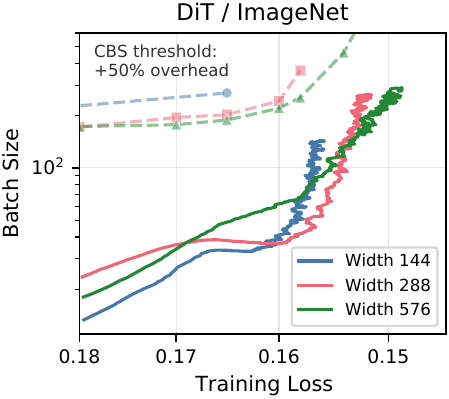}
    \includegraphics[width=0.32\linewidth]{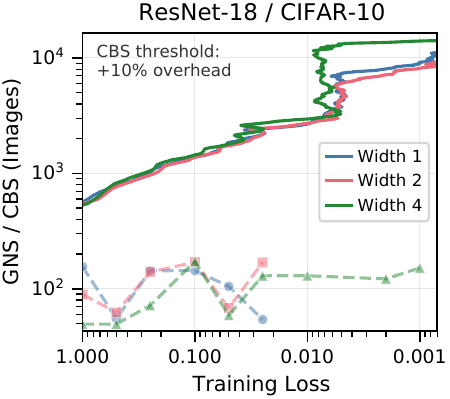}
    \includegraphics[width=0.32\linewidth]{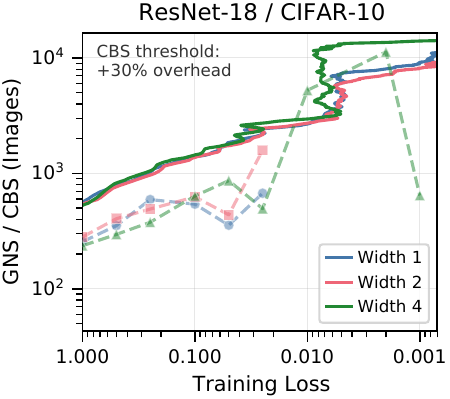}
    \includegraphics[width=0.32\linewidth]{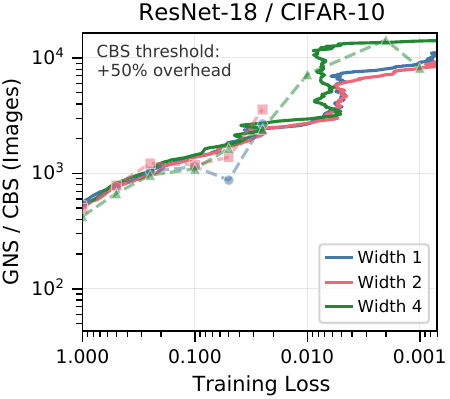}
    \caption{\ac{GNS} (solid lines) and \ac{CBS} (dashed, lighter lines) for
    tolerated overheads of $10\%$, $30\%$ and $50\%$ at the \ac{CBS} (left to
    right). Rows show GPT / FineWeb, DiT / ImageNet and ResNet-18 / CIFAR-10.
    The \ac{GNS} does not depend on the threshold, so the solid curves are
    identical within each row.}
    \label{fig:cbs_sensivity}
\end{figure}

Throughout the literature, different definitions of the \ac{CBS} exist, as discussed in Sec.~\ref{sec:crit_bs_definition}. In the following, we therefore recompute the \ac{CBS} under different overhead assumptions, complementing the $20\%$ threshold used in Sec.~\ref{sec:gns_cbs_loss}. As can be observed from the plots in Fig.~\ref{fig:cbs_sensivity}, a tighter definition of the \ac{CBS} (i.e., a lower tolerated overhead) moves the \ac{CBS} measurements closer to the \ac{GNS} predictions for the generative \ac{GPT} and \ac{DiT} models, although the \ac{GNS} still tends to underestimate the \ac{CBS} at all three thresholds. In the classification case, the \ac{GNS} continues to overestimate the \ac{CBS} at the $10\%$ and $30\%$ overhead thresholds (the tighter the threshold, the larger the overestimation) but aligns closely with the \ac{CBS} under the original $50\%$ definition of~\cite{mccandlish2018empirical}. The task-dependent deviation reported in Sec.~\ref{sec:gns_cbs_loss} thus never changes sign across thresholds, but its magnitude depends on the tolerated overhead.

\section{Sensitivity to Random Seed}
\label{sec:random_seeds}

\begin{figure}
    \centering
    \includegraphics[width=0.5\linewidth]{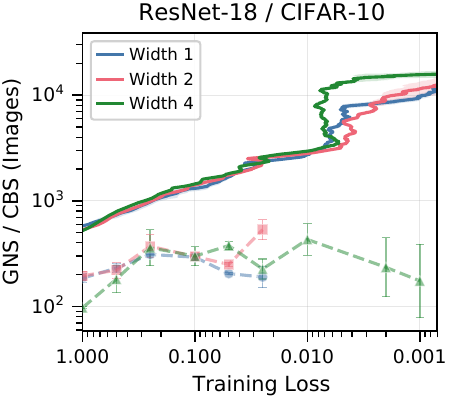}
    \caption{\ac{CBS} and \ac{GNS} for \ac{CV}, averaged over two seeds. The shaded
band (\ac{GNS}) and error bars (\ac{CBS}) span the range of the two seeds.}
    \label{fig:cbs_gns_cv_err}
\end{figure}

As the computations required to deliver the results proposed in this dataset are
expensive and scale linearly with the number of random seeds, mostly only a
single random seed is provided. However, for the \ac{CV} case, we reran the
batch size--learning rate grid from Tab.~\ref{tab:hyperparameters} for an additional seed to
compute error bars. We average the training loss and the \ac{GNS} of the two
seeds step by step before smoothing, and we repeat the \ac{CBS} extraction
separately for each seed. In Fig.~\ref{fig:cbs_gns_cv_err}, lines show the
seed-average, while the shaded band and the error bars show the range covered by
the two seeds. With only two seeds we report this range instead of a standard
error. The \ac{GNS} is almost identical for both seeds, whereas the \ac{CBS}
varies by a small factor, with the largest deviations at the
lowest target losses, but still tending to be overestimated by the \ac{GNS}.

\end{document}